%% file: main-5597-Yang.tex
\documentclass[11pt,a4paper]{article}
\usepackage{times,latexsym}
\usepackage{url}

\usepackage{amsmath}
\usepackage{amsfonts}
\usepackage{amssymb}
\usepackage{pifont}
\usepackage{bbm}
\usepackage{caption}
\usepackage{subcaption}
\usepackage{graphicx}
\usepackage{multirow}
\usepackage{multicol}
\usepackage[para]{threeparttable}
\usepackage{booktabs}
\usepackage{hyperref}
\usepackage{array}
\usepackage{colortbl}   
\usepackage{xspace}
\usepackage[normalem]{ulem}
\useunder{\uline}{\ul}{}
\usepackage{pgffor}
\usepackage{graphicx}
\usepackage{cleveref}

\usepackage{xparse}

\usepackage[T1]{fontenc}

\usepackage[acceptedWithA]{tacl2021v1}

\usepackage{xspace,mfirstuc,tabulary}

\newif\iftaclinstructions
\taclinstructionsfalse 
\iftaclinstructions
\renewcommand{\confidential}{}
\renewcommand{\anonsubtext}{(No author info supplied here, for consistency with
TACL-submission anonymization requirements)}
\newcommand{\instr}
\fi

\newif\ifdebugmode

\newcommand{\revised}[1]{%
  \ifdebugmode
    \textcolor{orange}{#1}%
  \else
    #1%
  \fi
}

\iftaclpubformat 

\else

\fi

\DeclareMathOperator*{\argmax}{arg\,max}
\DeclareMathOperator*{\argmin}{arg\,min}

\newcommand{\cmark}{\ding{51}}%
\newcommand{\xmark}{\ding{55}}%

\newcommand{\prob}{$p(\mathbf{y}|x,t)$\xspace} 

\NewDocumentCommand{\mn}{ m o o }{%
  \textnormal{#1}%
  \IfValueT{#2}{$_{\textnormal{#2}}$}%
  \IfValueT{#3}{$^{\textnormal{#3}}$}%
}

\newcommand{\mia}{\mn{MI}[A]\xspace}
\newcommand{\miag}{\mn{MI}[AG]\xspace}
\newcommand{\mial}{\mn{MI}[AL]\xspace}
\newcommand{\miagl}{\mn{MI}[AGL]\xspace}
\newcommand{\mdlm}{\mn{MDL}[M]\xspace}
\newcommand{\gem}{\mn{GE}[M]\xspace}

\newcommand{\pss}[1]{$\text{PSS}_{\text{#1}}$\xspace}
\newcommand{\qq}{$\tilde{q}(\mathbf{y}|x,t)$\xspace}
\newcommand{\pmidc}{PMI$_\text{DC}$\xspace}
\newcommand{\mic}{MI$_\text{A}^\text{(PA)}$}
\newcommand{\lec}{MDL$_\text{M}^\text{(PA)}$}

\makeatletter
\def\gcmidrule{\arrayrulecolor{gray!20}
    \noalign{\ifnum0=`}\fi
    \@ifnextchar[{\@gcmidrule}{\@gcmidrule[\cmidrulewidth]}}
\def\@gcmidrule[#1]{\@ifnextchar({\@@gcmidrule[#1]}{\@@gcmidrule[#1]()}}
\def\@@gcmidrule[#1](#2)#3{\@@@gcmidrule[#3]{#1}{#2}}
\def\@@@gcmidrule[#1-#2]#3#4{\global\@cmidla#1\relax
    \global\advance\@cmidla\m@ne
    \ifnum\@cmidla>0\global\let\@gtempa\@cmidrulea\else
    \global\let\@gtempa\@cmidruleb\fi
    \global\@cmidlb#2\relax
    \global\advance\@cmidlb-\@cmidla
    \global\@thisrulewidth=#3
    \@setrulekerning{#4}
    \ifnum\@lastruleclass=\z@\vskip \aboverulesep\fi
    \ifnum0=`{\fi}\@gtempa
    \noalign{\ifnum0=`}\fi\futurenonspacelet\@tempa\@xgcmidrule}
\def\@xgcmidrule{%
   \ifx\@tempa\gcmidrule
       \vskip-\@thisrulewidth
       \global\@lastruleclass=\@ne
   \else \ifx\@tempa\morecmidrules
       \vskip \cmidrulesep
       \global\@lastruleclass=\@ne\else
       \vskip \belowrulesep
       \global\@lastruleclass=\z@
   \fi\fi
   \ifnum0=`{\fi}
  \arrayrulecolor{black}}
\makeatother

\title{Improving Probability-based Prompt Selection \\ Through Unified Evaluation and Analysis}

\author{Sohee Yang{\textsuperscript{1$*$}} \quad Jonghyeon Kim{\textsuperscript{3$\dagger$}} \quad Joel Jang{\textsuperscript{4}} \\ {\bf Seonghyeon Ye{\textsuperscript{2}} \quad  Hyunji Lee{\textsuperscript{2}} \quad Minjoon Seo{\textsuperscript{2}}} \\
{\textsuperscript{1}}UCL\quad{\textsuperscript{2}}KAIST\quad{\textsuperscript{3}}Dongguk University\quad{\textsuperscript{4}}University of Washington \\ 
\texttt{sohee.yang.22@ucl.ac.uk}
}

\date{}

\begin{document}
\maketitle
\def\thefootnote{*}\footnotetext{This project was initiated while the first author was a Master's student at KAIST (Nov 2022 - Feb 2023).}\def\thefootnote{\arabic{footnote}}
\def\thefootnote{$\dagger$}\footnotetext{Work done as an intern at KAIST.}\def\thefootnote{\arabic{footnote}}

\begin{abstract}
Previous works in prompt engineering for large language models have introduced different gradient-free probability-based prompt selection methods that aim to choose the optimal prompt among the candidates for a given task but have failed to provide a comprehensive and fair comparison between each other.
In this paper, we propose a unified framework to interpret and evaluate the existing probability-based prompt selection methods by performing extensive experiments on 13 common and diverse NLP tasks. We find that each of the existing methods can be interpreted as some variant of the method that maximizes mutual information between the input and the predicted output (MI). Utilizing this finding, we develop several other combinatorial variants of MI and increase the effectiveness of the oracle prompt selection method from 87.79\% to 94.98\%, measured as the ratio of the performance of the selected prompt to that of the optimal oracle prompt. Furthermore, considering that all the methods rely on the output probability distribution of the model that might be biased, we propose a novel calibration method called Calibration by Marginalization (CBM) that is orthogonal to the existing methods and helps increase the prompt selection effectiveness of the best method to 96.85\%, achieving 99.44\% of the oracle prompt F1 without calibration.\looseness-1\footnote{The code and datasets used in our work are available at \href{https://github.com/soheeyang/unified-prompt-selection}{https://github.com/soheeyang/unified-prompt-selection}.}
\end{abstract}

\section{Introduction}

Large Language Models (LLMs) have demonstrated remarkable performance in solving various natural language processing tasks through prompt-based learning without requiring additional task-specific training~\citep{brown2020language,dong2023survey}. \revised{However, the performance of LLMs can heavily fluctuate according to the choice of prompts~\citep{Zhao2021-lq,Holtzman2021-wj,Lu2022-yp}. While various prompt engineering approaches have been proposed to mitigate this issue, the nontrivial prerequisites of many of these methods, such as training an additional model and/or using an additional component, have been a bottleneck to their real application~\citep{Liu2021-hv,li2021prefix,Jiang2020-rv,Prasad2022-qr,Liu2022-nx,Rubin2022-jx}.}\looseness-1

On the other hand, probability-based prompt selection methods do not require any additional parameter updates or additional components\footnote{While the prerequisite is a set of candidate prompts to select from, this data is relatively small in size and can be easily obtained from the research community~\citep{bach2022promptsource} or via machine generation~\citep{openai2023gpt4}.} and thus provide a promising and easily applicable solution; these methods aim to select the prompt from a set of prompts that is expected to be most effective in helping a language model to make correct predictions \textit{solely based on the probability distribution} of the model~\citep{Sorensen2022-nm,Lu2022-yp,Wu2022-hr,Liao2022-fe,Gonen2022-zf}.
However, despite their ease of utilization, there has been a lack of comprehensive comparative evaluation between existing probability-based prompt selection methods, as each method is proposed in different setups and evaluated on different datasets, evaluation instances, sets of prompts, and models.

\begin{figure}[t!]
\centering
\begin{subfigure}[b]{0.48\textwidth}
    \includegraphics[width=\textwidth]{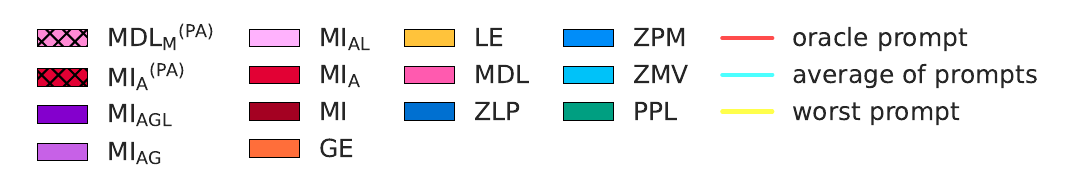}
    \end{subfigure}
    \begin{subfigure}[b]{0.192\textwidth}
    \includegraphics[width=\textwidth]{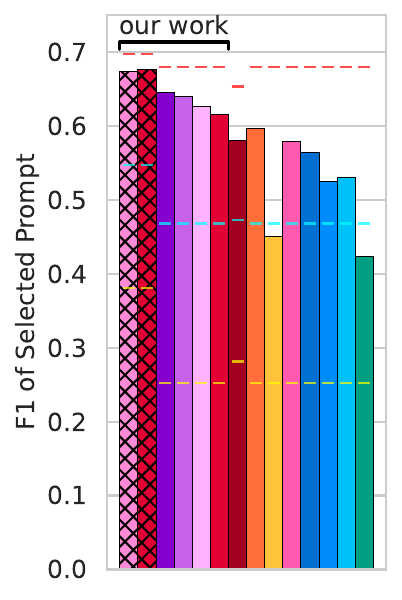}
    \caption{Prompt Selection}\label{fig:main_prompt} 
    \end{subfigure}
    \begin{subfigure}[b]{0.275\textwidth}
    \includegraphics[width=\textwidth]{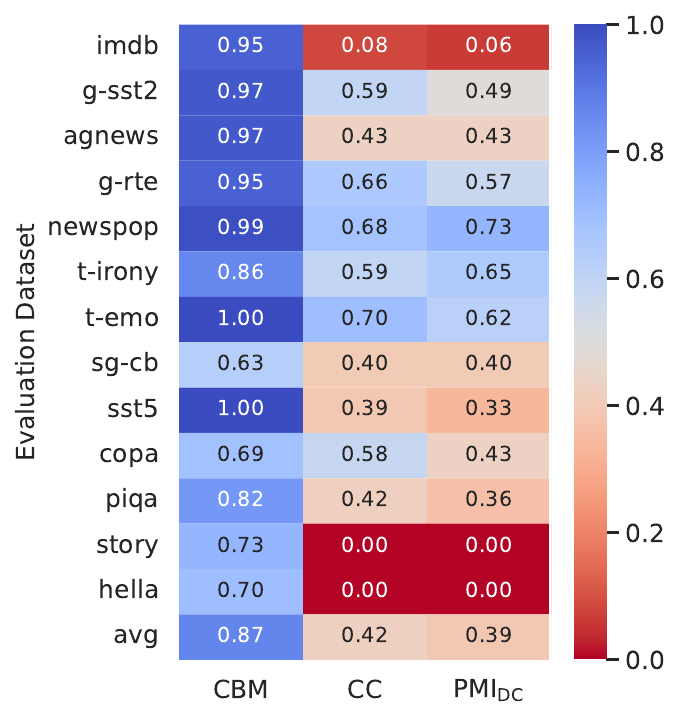}
    \caption{Answer Selection}\label{fig:main_answer} 
    \end{subfigure}
\caption{\textbf{(a)} F1 of the prompts selected by different probability-based prompt selection methods, averaged across 13 datasets. Per-dataset F1 and accuracy are shown in Figure~\ref{fig:granular}. The methods without super/subscripts are the existing methods (Table~\ref{tab:methods}), while those with super/subscripts are our proposed methods (Table~\ref{tab:transfer} \& Equation~\ref{eq:cbm}). \textbf{(b)} Ratio of the prompts (out of 100) whose F1 on each dataset improves by applying probability calibration for answer selection, averaged across 10 models. Our proposed calibration method, CBM (Equation~\ref{eq:cbm}), is considerably more effective than CC and \pmidc (Table~\ref{tab:calibration}) in enhancing the answer selection performance of the prompts.}\label{fig:main}
\end{figure}

In this paper, we first carefully design a unified evaluation setup to facilitate a fair comparison between different prompt selection methods. Our unified evaluation reveals that no single method consistently outperforms other methods across all datasets and that all existing probability-based prompt selection methods roughly correspond to a sub-term of the equation of Mutual Information (MI)~\citep{Sorensen2022-nm}. We utilize this discovery to propose several variants of MI that use different combinations of the components of existing methods, and the best combinational variant \miagl increases the scaled F1 (F1 divided by that of the oracle prompt, showing the effectiveness of the prompt selection method) from 87.79\% to 94.98\% (\miagl of Figure~\ref{fig:main_prompt}).

Furthermore, we find the need for a better approximation of the LLM's output probability distribution, considering that all probability-based prompt selection methods rely on the probabilistic estimates from the model that might be biased. Therefore, by drawing a connection between the existing model output probability calibration methods~\citep{Zhao2021-lq,Holtzman2021-wj}, we propose an enhanced calibration method, Calibration By Marginalization (CBM). CBM significantly improves the prompt selection performance of several methods when applied to calibrate the output probability of LLMs, increasing the best-scaled F1 to 96.85\% (\mic of Figure~\ref{fig:main_prompt}), achieving 99.44\% of the oracle prompt F1 under the uncalibrated scenario. CBM also proves to show the most robust answer selection enhancement across multiple datasets compared to the existing calibration methods (Figure~\ref{fig:main_answer}).

\section{Probability-based Prompt Selection}

In this section, we perform a unified evaluation of existing probability-based prompt selection methods. First, we describe the task of probability-based prompt selection in Section~\ref{sec:task}. Next, we briefly introduce each of the existing methods in Section~\ref{sec:overview}. Then, we describe our experimental setup for unified evaluation in Section~\ref{sec:unified_setup} and present the evaluation results in Section~\ref{sec:unified_eval}.

\subsection{Task Description}\label{sec:task}
\input{tables/approaches}
Probability-based prompt selection is the task of selecting one or more prompts from a list of prompts $T$, which are expected to help the language model $\theta$ make the most accurate prediction for the evaluation dataset $X$ where the evaluation instances are drawn from the data distribution, $x \sim P_X$, utilizing only the output probability distributions of the model on $X$,\footnote{Note that one can perform a computation-efficient prompt selection or transfer of prompt selection by (1) selecting \textit{one} prompt using a \textit{subset of $X$} or a \textit{separate development set $X'$} and then (2) use the selected prompt for the target evaluation dataset $X$ to instantiate all $x \sim P_X$. However, following the conventional setup of the previous works and for comparison with instance-wise prompt selection methods where such an approach is not applicable by design, we do not use a separate $X'$.} without knowing the ground truth labels and using neither additional gradient-based updates nor other trained components.
The performance of a probability-based prompt selection method is evaluated by how high the score of the evaluation metric obtained with the selected prompt(s) is.

When one prompt is selected for the whole dataset, the performance is upper bounded by the performance obtained with the prompt with which the model achieves the best metric score; we call such a prompt the optimal oracle prompt.\footnote{The number of oracle prompts can be greater than one, but we use the singular form for a more concise presentation.} When one prompt is selected for each $x \sim P_X$, different $t \in T$ can be chosen for each $x$; we call such a prompt selection approach as instance-wise prompt selection.

Note that the definition of prompt can vary according to the setup for which prompt selection is performed. When prompt selection is applied to zero-shot learning, prompts are defined as various formats of \textit{text templates} that are filled by evaluation instances $x \sim P_X$ to facilitate. On the other hand, for few-shot (in-context) learning, prompts are often defined as the \textit{demonstrations} sampled from a training/development set or texts of permutations of such demonstrations. In our work, in order to enable comparison between all the methods proposed either in zero-shot and few-shot setup, we perform prompt selection in a zero-shot setup with the former definition of prompt.\footnote{We have performed additional experiments in a few-shot learning setup using the texts of permutations of varying numbers of in-context learning demonstrations as the prompts. However, we do not include these results in the paper due to space limitations; also, the overall trend of the results stays similar to that of the zero-shot learning setup.}

\paragraph{Concrete Example}
Examples of prompts $t \in T$ include ``Which category does the following news article fall into? \{text\}'', ``The following news article, \{text\}, covers the topic of'', and ``\{text\} belongs in which category: Politics, Sports, Business, Science and Technology''. We say that $x$ instantiates the prompt $t$ when $x$ is inserted into the placeholder \{text\} of the prompt template and let $\iota(x,t)$ denote the instantiated prompt. Each of the answer categories represents the concept of politics, sports, business, and science/technology, and uses ``Politics,'' ``Sports,'' ``Business,'' and ``Science and Technology'' as the verbalizer (the actual text evaluated to score the answer choices), respectively.

For instance, given OPT 2.7B~\citep{zhang2022opt} as the language model, ``King Charles III's Coronation watched by more than 18 million viewers'' as $x$, and the three prompts shown as examples in the previous paragraph, a prompt selection method should choose the prompt that is most likely to help OPT 2.7B correctly predict the answer $y$ among the possible answer choices $Y$ which represent the concepts of politics, sports, business, and science/technology. To select such a prompt, the method must rely solely on the output probability of the model given the instantiated prompts as input, e.g., $p(\text{``Politics''}|\text{``Which category \dots\ King \dots''})$.

\subsection{Existing Approaches}\label{sec:overview}

Table~\ref{tab:methods} provides the summary of the existing approaches for probability-based prompt selection. In the equations, we use $p(\mathbf{y}|x,t) \in \mathbb{R}^{|Y|}$ to express the output probability distribution of the model over the answer choices, $P_\theta(Y|X=x,T=t)$, when the instantiated prompt $\iota(x,t)$ is given as the input. The probability for each $y \in Y$ is calculated as
\begin{equation*}
    p(y|x,t) = \frac{\exp (\log \tilde{p}(y|x,t))}{\sum_{y' \in Y} \exp (\log \tilde{p}(y'|x,t))},
\end{equation*}
\noindent
where $\log \tilde{p}(y|x,t)$ is the unnormalized logit that the model outputs.
When $y$'s verbalizer is tokenized into more than one token, we calculate $\log \tilde{p}(y|x,t)$ as the mean of log-probability over the tokens of the verbalizer for datasets with fixed answer choices, and as the sum of log-probability for datasets with dynamically changing sentence-type answer choices, except for the method proposed by \citet{Sorensen2022-nm} which explicitly specifies that the calculation of \prob uses only the logits of the first token (dubbed as One-Token Response (OTR) in their work).
We use $\text{H}(q(\mathbf{y}))$ to denote the entropy of an arbitrary probability distribution $q(\mathbf{y}) \in \mathbb{R}^{|Y|}$, $-\sum_{y \in Y} q(y) \log q(y)$. When $q(\mathbf{y}) = p(\mathbf{y}|x,t)$, we use $\text{H}(Y|x,t)$ to represent its entropy $\text{H}(Y|X=x,T=t)$.

\paragraph{Mutual Information (MI)} \citet{Sorensen2022-nm} propose to select one prompt for the evaluation dataset that maximizes the mutual information between the evaluation instances $X$ and their corresponding model predictions $Y$ given prompt $t$, $\text{I}(Y;X|t) = \left[ \text{H}(Y|t) - \text{H}(Y|X, t) \right]$. Since they use the assumption that $p(x|t) = P_X(X=x) = \frac{1}{|X|}$, the equation becomes as shown in the first row of Table~\ref{tab:methods}. The intuition of the method is to select the prompt that guides the model to make less biased predictions on average (high $\text{H}(Y|t)$) and confident predictions about the input data (low \text{H}(Y|X, t)).

\paragraph{Entropy (GE, LE)} \citet{Lu2022-yp} propose to select the prompt (finding the best ordering of few-shot demonstrations for in-context learning in their setup) using entropy-based metrics. While their proposed methods are intended specifically for in-context learning, viewing prompts as texts of permutations of demonstrations\footnote{They generate a probing set with demonstrations from the training set and use the probing set to find the best order.}, we adopt the methods for our zero-shot setup of selecting among text template prompts and thus do not use an additional training set or construct a probing set. Global Entropy (GE) or Local Entropy (LE) shown in the second row of Table~\ref{tab:methods} are used to select a single prompt among the prompt candidates for the evaluation dataset.

\paragraph{Minimum Description Length (MDL)} \citet{Wu2022-hr} propose to select the prompt (a permutation of few-shot demonstrations in their setup) that requires minimum codelength to compress and transmit testing label $y$ given the testing input $x$. With several assumptions and approximations presented in Section 4.3 of the work of \citet{Wu2022-hr}, the equation boils down to finding different $t$ for each $x \in X$, $\argmin_t \text{H}(Y|x,t)$, performing instance-wise prompt selection.
As their original setup for prompt selection is few-shot learning, they perform demonstration sampling as a set selection and then rank the texts of different permutations of the demonstrations. Here, we describe only the ranking part of their approach that we employ for our zero-shot learning setup.

\paragraph{Zero-Label Prompt Selection (ZLP, ZPM, ZMV)} \citet{Liao2022-fe} propose to make a pseudo-label for each $x$ by ensembling the outputs for all prompts to make a score $s(x,y)$ for each $x$, and then choosing one prompt $t$ for the evaluation dataset whose cases of $\argmax_{y \in Y}p(y|x,t) = \argmax_{y \in Y}s(x,y)$ is the maximum. As shown in Table~\ref{tab:methods}, they propose three ways to calculate $s(x,y)$: using the ensemble of log-probability mean, probability mean, and majority vote. We refer to them as ZLP, ZPM, and ZMV, respectively. While the authors of the original work applied filtering of prompts, we observed from our preliminary experiments that filtering does not have a significant effect.

\paragraph{Perplexity (PPL)} \citet{Gonen2022-zf} propose to select one prompt for the evaluation dataset with which the language model exhibits the lowest average perplexity of the instantiated prompt $\iota(x,t)$ as shown in the last row of Table~\ref{tab:methods}. $p(x,t)$ is calculated as $\left[\Pi_{i=1}^{|\iota(x,t)|} p(\iota(x,t)_i | \iota(x,t)_{<i})\right]^{\frac{1}{|\iota(x,t)|}}$, where $\iota(x,t)_i$ represents the $i$-th token of the instantiated prompt $\iota(x,t)$. We include the geometric mean to the definition of $p(x,t)$ because the averaged probability is often used to approximate the probability of a sequence.

\subsection{Experimental Setup}\label{sec:unified_setup}

\paragraph{Evaluation Datasets}
Our dataset selection, aimed at fair measurement of various probability-based prompt selection methods, is guided by several factors. We favor the datasets previously used in research, those encompassing diverse domains, and datasets where prompt selection is meaningful. We exclude the datasets where all prompts underperform a random baseline or where a naive baseline of selecting the mode label could excel due to high imbalance. By excluding the datasets with high imbalance, we aim to avoid the false positive cases where a failed algorithm that collapses to select one label regardless of the input is evaluated as a competitive method by chance.

\input{tables/dataset}

The selected datasets have diverse label types and distributions, and we categorize them based on their label distributions into balanced (label distribution is about 1:1), unbalanced (otherwise), and dynamic\footnote{The answer choices are sentences and vary dynamically for each evaluation instance. In these datasets, the label index is not connected to some concept, unlike the datasets with static choices (e.g., 0 is negative and 1 is positive in sst2), so the ratio of labels is not meaningful. However, all the datasets of dynamic categories that we use have balanced label distribution.} categories. The 13 datasets selected through this process are shown in Table~\ref{tab:dataset}.\footnote{\citet{imdb,wang2019glue,Zhang2015CharacterlevelCN,Moniz2018MultiSourceSF,barbieri2020tweeteval,mohammad2018semeval,van2018semeval,wang2019superglue,socher2013recursive,Bisk2020,mostafazadeh2017lsdsem,zellers2019hellaswag}}

\paragraph{Prompts}
We create a diverse range of 100 prompts for each of the 13 evaluation datasets, which results in 1,300 prompts in total. For each dataset, a few of the 100 prompts are taken from PromptSource~\citep{bach2022promptsource}, and the rest are generated using GPT 3.5~\citep{openai2023gpt4} to speed up the prompt generation process and then manually reviewed and corrected\footnote{The generation, review, and correction are done by the first two authors of the paper.}.  The prompts are designed to encompass various formats, with the evaluation instance and sometimes the answer choices appearing at different positions within the prompt, to ensure that the prompt selection task is meaningful. Table~\ref{tab:prompt-examples} shows a few examples of the prompts. We use one-token words as the verbalizers for the answer choices in most prompts, except for the prompts for the datasets of the dynamic category.\looseness-1

\paragraph{Models}
We conduct the majority of our experiments with ten different models of varying sizes ranging from 1.3B to 66B\footnote{GPT-Neo~\citep{gpt-neo} 1.3B, OPT~\citep{zhang2022opt} 1.3B, GPT2-XL~\citep{Radford2019LanguageMA}, GPT-Neo 2.7B, OPT 2.7B, BLOOM 3B~\citep{workshop2023bloom}, GPT-J 6B, OPT 6.7B, OPT 30B, and OPT 66B}. However, to present the experimental results and analysis more clearly, we only display the results of OPT 2.7B throughout the paper since \textit{the overall trend remains mostly identical} (shown in Section~\ref{sec:more_analysis}).
\begin{table}[t!]
\setlength{\tabcolsep}{4pt}
\resizebox{0.48\textwidth}{!}{
\begin{tabular}{lp{8cm}p{3cm}}
\toprule
\textbf{Dataset} & \textbf{Prompt} & \textbf{Verbalizers for $Y$} \\
\midrule
imdb  & From the following review, can you tell whether the sentiment is positive or negative? & negative, positive \\
agnews  & Which category among Politics, Sports, Business, Scienc would this news article fall under? & Politics, Sports, Business, Science \\
g-rte & Given the statement "\{\{sentence1\}\}", does it necessarily follow that "\{\{sentence2\}\}" is true? & yes, no \\
sg-cb & If the above statement is true, can we conclude that "\{\{hypothesis\}\}" is also true? Yes, no, or maybe? & Yes, no, maybe \\
sst5  & What is the sentiment expressed in the following sentence? It's either terrible or negative or neutral or positive or excellent. "\{\{ text \}\}" & terrible, negative, neutral, positive, excellent\\
piqa  & Your task is to achieve: \{\{goal\}\}\textbackslash{}n\textbackslash{}nWhich of the following options is the most appropriate?\textbackslash{}n\textbackslash{}n- \{\{sol1\}\}\textbackslash{}n- \{\{sol2\}\} \textbackslash{}n\textbackslash{}nAnswer: & \{\{sol1\}\}, \{\{sol2\}\} \\
\bottomrule
\end{tabular}}
\caption{Examples of the created prompts. The prompts are written in Jinja for the use of PromptSource~\citep{bach2022promptsource} APIs.}
\label{tab:prompt-examples}
\end{table}

\paragraph{Evaluation Metrics}
Prompt selection performance is assessed using macro F1 of the selected prompts. To compare the effectiveness of the prompt selection methods across different datasets or models, we normalize the value by the performance of the oracle prompt (upper bound) and present it as scaled F1.

\paragraph{Implementation Details}
We use a modified version of the codebase of \citet{sanh2022multitask}\footnote{\href{https://github.com/bigscience-workshop/t-zero}{https://github.com/bigscience-workshop/t-zero}} and PromptSource~\citep{bach2022promptsource}\footnote{\href{https://github.com/bigscience-workshop/promptsource}{https://github.com/bigscience-workshop/promptsource}} to run model inference and add custom prompts, respectively. The inference is performed using one to four NVIDIA V100 32GB GPUs.

\subsection{Experimental Results}\label{sec:unified_eval}
\begin{figure}[t!]
\centering
\begin{subfigure}[b]{0.45\textwidth}\includegraphics[width=\textwidth]{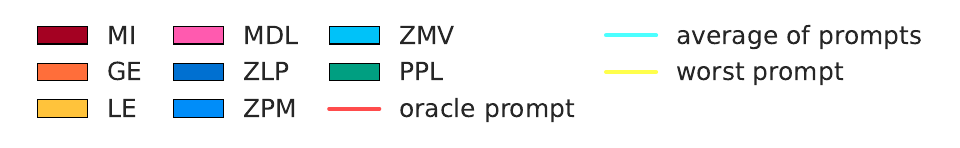}\end{subfigure}
    \begin{subfigure}[b]{0.45\textwidth}
    \includegraphics[width=\textwidth]{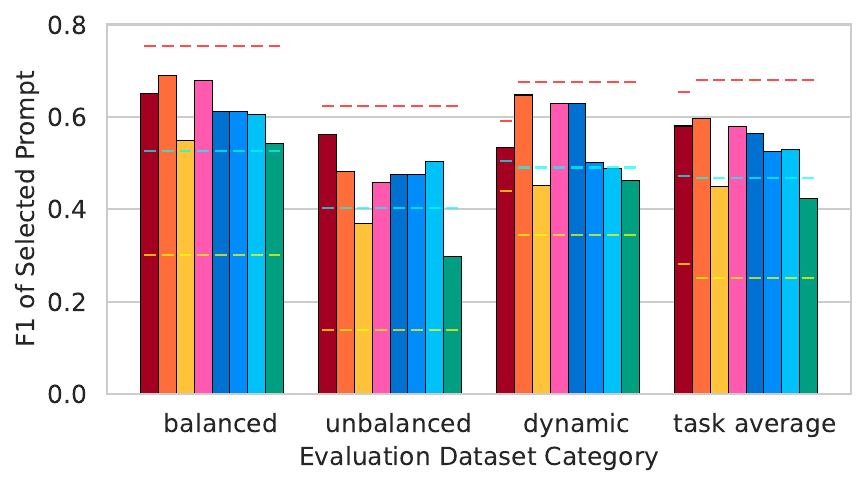}
    \end{subfigure}
\caption{F1 of the prompts selected by the existing probability-based prompt selection methods, averaged for each dataset category, with the task average also shown.}\label{fig:existing}
\end{figure}

We find that there is no single probability-based prompt selection method that consistently outperforms one another across all 13 datasets and evaluation categories. While PPL and LE do not rank first in any dataset, every other method ranks first in a few datasets.
Figure~\ref{fig:existing} illustrates the selected prompt performance averaged by category, along with the performance of the best (oracle) and worst prompts and the average performance of all prompts. In the balanced category, GE and MDL outperform others, with MI closely following. In the unbalanced category, MI stands out, while in the dynamic category, GE, MDL, and ZLP perform the best. LE and PPL generally underperform in all of the datasets; their task average does not even exceed the average performance of all prompts.\footnote{Interpretations of these results are provided in Section~\ref{sec:revisit}.} We conclude that no single existing approach is significantly better than others, especially when dividing the evaluation dimensions into balanced, unbalanced, and dynamic labels.
\section{Improving MI via Unified Analysis}\label{sec:unified_analysis}

In this section, we first derive a unified view of prompt selection methods in Section~\ref{sec:connections} and show that each method other than MI roughly corresponds to a sub-term of the equation of MI and revisit the previous experimental results for a unified analysis in Section~\ref{sec:revisit}. Then, from the unified view and analysis, we identify the differences between methods, particularly MI, GE, and MDL, and derive a few combinational variants by transferring design elements across methods which improves the prompt selection performance of MI.

\subsection{Unified View: Identifying Connections Between Methods}\label{sec:connections}

\begin{figure}[t!]
\centering
    \begin{subfigure}[b]{0.45\textwidth}
    \includegraphics[width=\textwidth]{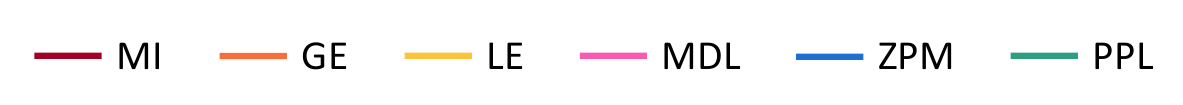}
    \end{subfigure}
    \begin{subfigure}[b]{0.45\textwidth}
    \includegraphics[width=\textwidth]{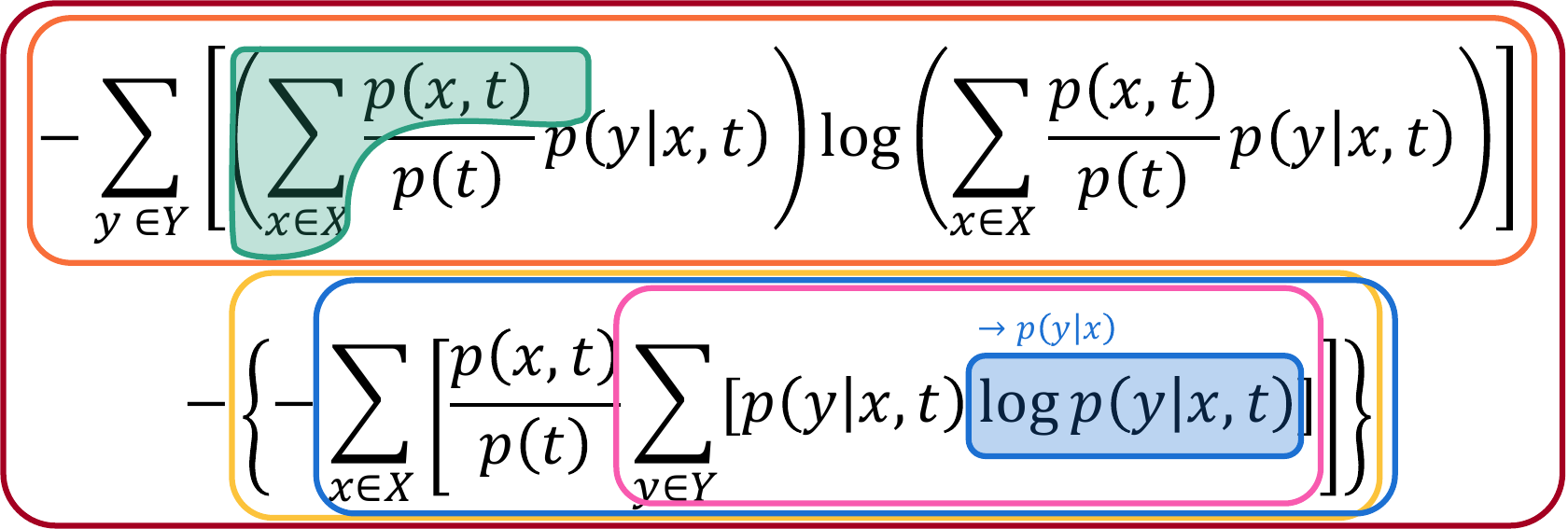}
    \end{subfigure}
\caption{The highlighted parts of the equation are rough estimations of the Prompt Selection Score (PSS) of each method, i.e., the score of which the prompt with the maximum value is chosen by the prompt selection method. They show the connection between different probability-based prompt selection methods.}\label{fig:unified}
\end{figure}

\paragraph{Prompt Selection Score (PSS)} Figure~\ref{fig:unified} offers a unified view of existing probability-based prompt selection methods, highlighting that each method except for MI approximately corresponds to a sub-term in the equation of MI. We denote the highlighted parts as the \underline{P}rompt \underline{S}election \underline{S}core of each method (\pss{method}); the score of which the prompt with the maximum value is chosen by the prompt selection method.

\paragraph{MI vs. GE and LE} MI selects a prompt that maximizes the first term of \pss{MI}, $\argmax_t \text{H}\left(\frac{1}{|X|} \sum_x p(\mathbf{y}|x,t)\right)$, and minimizes the second term, $\frac{1}{|X|} \sum_x \text{H}\left(Y|x,t\right)$. This means that MI favors prompts that provide balanced predictions without label bias (interpretation of the first term) and sharp answer prediction distribution across all instances in the dataset (interpretation of the second term). These terms roughly correspond to \pss{GE} and $-$\pss{LE}, respectively. The difference between \pss{GE} and the first term of \pss{MI} is that the former converts \prob to one-hot before taking the entropy of the mean. In sum, the prompts selected by GE and MI align, while those chosen by LE and MI tend to be opposite. Note that one expected caveat of GE is that it will be less effective when the dataset itself has a label imbalance.
\paragraph{MI vs. MDL} MDL is the only method among the presented probability-based prompt selection methods that selects a different prompt for each evaluation instance $x$, i.e., performs instance-wise prompt selection. Essentially, MDL is an instance-wise version of the second term of \pss{MI}, choosing prompts whose output probability distribution $p(\mathbf{y}|x,t)$ has the lowest entropy, and thus aligns with MI. Since MDL favors the prompt that makes the model output a sharp probability distribution, one expected caveat of MDL is that it will not work well when the model fails to solve the given task and collapses to a single prediction regardless of the input with overly high confidence.

\paragraph{MI vs. ZPM} Zero-label prompt selection methods ensemble the results of all prompts to calculate $s(x,y)$, create pseudo labels by converting $s(x,y)$ to one-hot, and then choose the prompt with predictions most similar to the pseudo labels. Applying this view to \pss{ZPM} with an assumption of $p(t|x) = \frac{1}{|T|}$ results in an alternative form,
{\fontsize{10.46}{10.46}
\setlength{\thinmuskip}{1mu}
\setlength{\medmuskip}{2mu}
\setlength{\thickmuskip}{3mu}
\begin{align*}
    \text{PSS}_\text{ZPM} =& \sum_{x \in X} \text{one\ hot} \left(p(\mathbf{y}|x,t)\right)^{\top} \text{one\ hot} \left(\mathbf{s}(x,y)\right)\\
    &\text{s.t.}\ \mathbf{s}(x,y) = \frac{1}{|T|} \sum_{t \in T} p(\mathbf{y}|x,t) \approx p(\mathbf{y}|x)\\
    \therefore \text{PSS}_\text{ZPM} \approx& \sum_{x \in X} \text{one\ hot} \left(p(\mathbf{y}|x,t)\right)^{\top} \text{one\ hot} \left(p(\mathbf{y}|x)\right) \\
    \approx& \frac{1}{|X|} \sum_{x \in X} p(\mathbf{y}|x,t)^{\top} \log p(\mathbf{y}|x),
\end{align*}
}%
\noindent which roughly corresponds to the negation of the second term of \pss{MI}, well-aligning the two methods.\footnote{One expected caveat of the methods of zero-label prompt selection is that it might not work well when a large portion of the prompts fail to solve the given task. Therefore, \citet{Liao2022-fe} propose a way to filter out low-quality prompts in advance, but the filtering algorithm does not benefit their proposed methods in our experimental setup.}

\paragraph{MI vs. PPL} \pss{PPL} is the most dissimilar from \pss{MI}, along with \pss{LE}. Since $\argmin_t \frac{1}{|X|} \sum_x \frac{1}{p(x,t)} = \argmax_t \sum_x p(x,t)$, \pss{PPL} can be expressed as $\sum_x p(x,t)$. It is clear that \pss{PPL} differs from \pss{MI} because it considers the probability of $x$ and $t$ that \pss{MI} neglects. Applying the probabilistic assumption of MI $\left(p(x|t) = p(x) = \frac{1}{|X|}\right)$ to \pss{PPL} converts the equation to $\sum_x \frac{p(t)}{|X|}$, causing PPL to select the prompt with the lowest perplexity irrespective of the input. Since \citet{Gonen2022-zf} even restrict their prompt format for the input $x$ to appear at the beginning so that $p(x,t)$ is calculated only as the form of $p(t|x)p(x)$, i.e., the probability of prompt is always conditioned on $x$, the probabilistic assumption of MI is incompatible with the motivation of PPL.\footnote{Note that our experimental setup also differs with the setup of \citet{Gonen2022-zf}; we generated the prompts in an unrestricted manner that $x$ can appear anywhere in the prompt.}

\subsection{Unified Analysis: Revisiting Experimental Results}\label{sec:revisit}

Revisiting the unified evaluation in Section~\ref{sec:unified_eval}, the results align with our analysis from Section~\ref{sec:connections}. GE performs well in balanced datasets but poorly in unbalanced ones due to its preference for prompts that create balanced predictions. GE also performs well in dynamic datasets since the label distribution is balanced by chance (Table~\ref{tab:dataset}). MDL performs comparably to GE due to similar entropy calculations. LE's performance, however, is less satisfactory, given its optimization contradicts MDL. The underperformance of PPL compared to that by \citet{Gonen2022-zf} might be due to our use of diverse prompt formats\footnote{We allow the input $x$ to appear anywhere in the prompt, unlike their restricted setup where $x$ always comes at the beginning.}.

Note that in dynamic datasets, MI's best, worst, and average prompt performances differ due to its distinct calculation of \prob that uses only the first token logits; for other methods, \prob is calculated using all tokens (Section~\ref{sec:overview}).\footnote{In balanced and unbalanced cases, the number of tokens of most verbalizers is 1, so the best, worst, and average prompt performances of the prompts whose performance is calculated using only the first token are identical to the other methods; on the other hand, the verbalizer is a sentence for dynamic datasets and makes the difference.} This leads to a question: \textit{Is the difference in the calculation of \prob the reason that MI performs well in balanced and unbalanced cases but poorly in dynamic cases?} In addition, despite GE and MDL maximizing MI's sub-term, they outperform MI in balanced datasets. This observation leads to another question: \textit{Is their higher performance due to their one-hot \prob and instance-wise prompt selection?}

In the following subsection, we show that the answers to both questions are \textit{yes}, demonstrating that using all tokens to calculate \prob, one-hot \prob, and instance-wise prompt selection improves the prompt selection performance of MI.

\subsection{Experimental Results: Transferring Design Choices from Unified Analysis}\label{sec:transfer}

\paragraph{\prob calculation using all tokens helps MI.}
\begin{figure}[t!]
\centering
    \begin{subfigure}[b]{0.48\textwidth}
    \includegraphics[width=\textwidth]{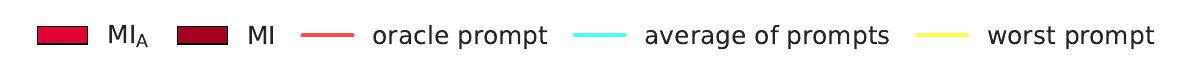}
    \end{subfigure}
    \begin{subfigure}[b]{0.48\textwidth}
    \includegraphics[width=\textwidth]{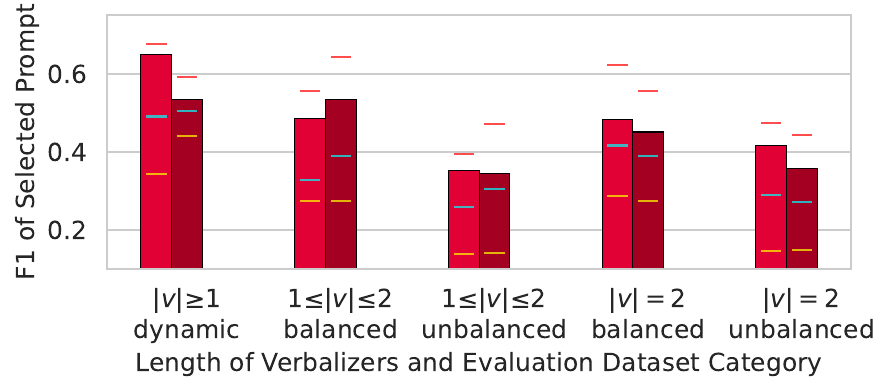}
    \end{subfigure}
\caption{F1 of the prompts selected by \mia and MI, averaged for each setup of a different number of tokens of verbalizers and evaluation dataset category. $|v|$ denotes the number of tokens of the verbalizers.}\label{fig:token}
\end{figure}
\begin{table}[t!]
\centering
\resizebox{0.48\textwidth}{!}{
\begin{tabular}{llcccl}
\toprule
 & & \textbf{A} & \textbf{G} & \textbf{L} & 
\textbf{Prompt Selection Score} \\ 
\midrule
\multicolumn{5}{l}{\textbf{Existing Methods}} \\\gcmidrule(lr){2-6}
&GE & \cmark & \cmark & - & $\text{H}\left(\frac{1}{|X|} \sum_x \text{one\ hot}(p(\mathbf{y}|x,t))\right)$ \\\gcmidrule(lr){2-6}
&MDL & \cmark & - & \cmark & $-\text{H}(Y|x,t)$ \\\gcmidrule(lr){2-6}
&MI & \xmark & \xmark & \xmark & $\text{GE}_\text{M} + \text{MDL}_\text{M}$ \\
\midrule
\multicolumn{5}{l}{\textbf{Explored Variants}} \\\gcmidrule(lr){2-6}
&GE$_\text{M}$ & \cmark& \xmark & - & $\text{H}\left(\frac{1}{|X|} \sum_x p(\mathbf{y}|x,t)\right)$ \\\gcmidrule(lr){2-6}
&MDL$_\text{M}$ & \cmark& - & \xmark & $-\frac{1}{|X|} \sum_x \text{H}\left(Y|x,t\right)$ \\\gcmidrule(lr){2-6}
&MI$_\text{A}$ & \cmark & \xmark & \xmark & $\text{GE}_\text{M} + \text{MDL}_\text{M}$ \\\gcmidrule(lr){2-6}
&MI$_\text{AG}$ & \cmark& \cmark & \xmark & $\text{GE} + \text{MDL}_\text{M}$ \\\gcmidrule(lr){2-6}
&MI$_\text{AL}$ & \cmark& \xmark & \cmark & $\text{GE}_\text{M} + \text{MDL}$ \\\gcmidrule(lr){2-6}
&MI$_\text{AGL}$ & \cmark& \cmark & \cmark & $\text{GE} + \text{MDL}$ \\
\bottomrule
\end{tabular}
}
\caption{\textbf{Top}: differences among GE, MDL, and MI. \textbf{Bottom}: new variations created by transferring design choices from existing probability-based prompt selection methods. \textbf{A} represents $p(\mathbf{y}|x,t)$ using \underline{A}ll tokens, \textbf{G} represents one-hot \prob like \underline{G}E, and \textbf{L} represents instance-wise selection (select for each $x$) like MD\underline{L}.}\label{tab:transfer}
\end{table}

To investigate the difference between using only the first token probability and the mean/sum of all tokens to calculate \pss{MI}, we develop a variant of MI called \mia (A of \underline{A}ll). Unlike MI and like other methods, \mia calculates \prob by taking the mean of all token logits for balanced and unbalanced datasets, and the sum for dynamic datasets. Since the balanced and unbalanced datasets in our experimental setup (Section~\ref{sec:unified_eval}) mostly use one-token verbalizers which result in the same result of MI and \mia, we utilize new sets of verbalizers of 1-2 tokens ($1\le|v|\le2$) or 2 tokens ($|v|=2$) for all the prompts of our evaluation datasets and compare the two methods. Our results in Figure~\ref{fig:token} show that using all tokens is more effective in all configurations except for the 1-2 token-balanced tasks.

\paragraph{One-hot \prob and instance-wise prompt selection benefits MI.}

We create combinational variants of GE, MDL, and MI (outlined in Table~\ref{tab:transfer}) to study whether their differences contribute to MI's lower performance in balanced datasets. For instance, \pss{\gem} is an \underline{M}I-like version of GE employing \prob without one-hot encoding, while \pss{\mdlm} is an \underline{M}I-like MDL version using the average of $\text{H}(Y|x,t)$ for all $x$ to select a single prompt. Contrarily, \miag and \mial are variants of MI, with the former emulating GE and the latter mirroring MDL, on top of \mia. \miagl is another MI variant employing the sum of \pss{GE} and \pss{MDL} as PSS, using one-hot \prob for the first term calculation and instance-wise selection.
\begin{figure}[t!]
\centering
    \begin{subfigure}[b]{0.48\textwidth}
    \includegraphics[width=\textwidth]{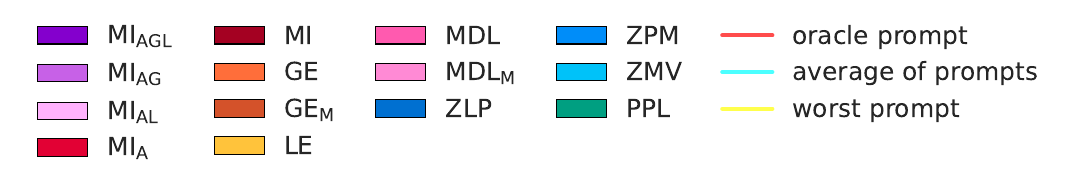}
    \end{subfigure}
    \begin{subfigure}[b]{0.48\textwidth}
    \includegraphics[width=\textwidth]{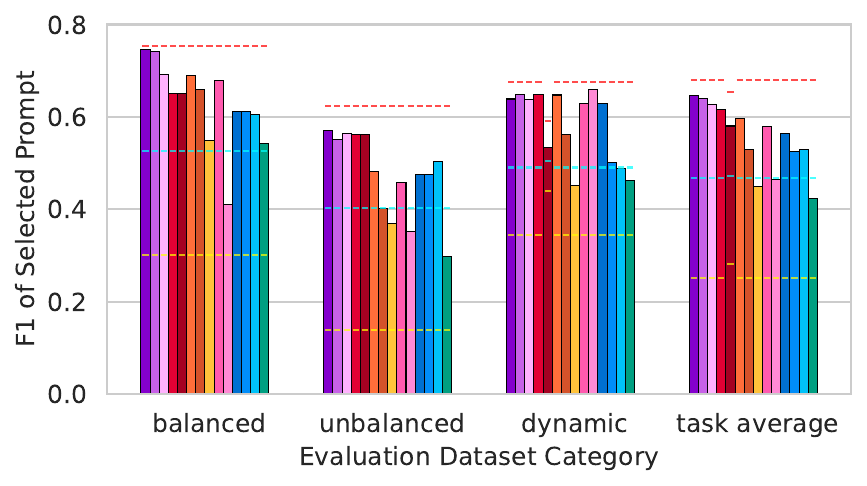}
    \end{subfigure}
\caption{F1 of the prompts selected by different probability-based prompt selection methods, averaged for each dataset category, with the task average also shown. The methods with subscripts are the combinational variants proposed in this subsection, whose Prompt Selection Scores are shown in Table~\ref{tab:transfer}\label{fig:transfer} . The methods with subscript M are combinational variants that use the component of \underline{M}I; the methods with L perform instance-wise prompt selection like MD\underline{L}; the methods with G utilize one-hot \prob like \underline{G}E. The methods with A use \underline{A}ll tokens to calculate \prob.}
\end{figure}
Figure~\ref{fig:transfer} compares these variants with existing methods. The variants that use instance-wise prompt selection (\miagl, \mial, MDL) perform better in balanced and unbalanced datasets but underperform in dynamic ones. Particularly in balanced datasets, \miagl, \mial, and \mia show significant improvement. While no method is consistently superior across all datasets (as observed in Section~\ref{sec:unified_eval}), \miagl significantly improves scaled F1 to 94.98\% (0.6454/0.6795) compared to that of the best existing method (GE), which is 87.79\% (0.5965/0.6795).

\section{Improving Prompt Selection Through Enhanced Probability Calibration}

While the previous section enhances prompt selection performance using combinatorial variants, in this section, we explore an orthogonal approach to further improve prompt selection: model output probability calibration.

Since all the prompt selection methods except for PPL depend on the model output probability \prob to calculate Prompt Selection Score (PSS), the stability and reliability of \prob affect their prompt selection performance. However, previous works have pointed out that \prob is unstable without calibration.\footnote{\citet{Zhao2021-lq} find that the probability in few-shot learning tends to favor certain answer choices appearing at the end of the prompt or common in pretraining data. \citet{Holtzman2021-wj} note that ranking based on string probability can be probabilistic due to surface form competition.}
To address the issue, \citet{Zhao2021-lq} suggest Contextual Calibration (CC), which reduces bias towards each answer choice by employing content-free inputs (``N/A'', ``[MASK]'', ``''), while \citet{Holtzman2021-wj} present Domain Conditional Pointwise Mutual Information (PMI$_\text{DC}$) by reweighting each answer choice based on its task-specific prior likelihood. We summarize the two methods for answer selection in Table~\ref{tab:calibration}; $\argmax_y$ \qq is selected as the answer, where \qq is the calibrated score.

One might assume that these existing calibration methods would effectively calibrate \prob for PSS. However, through the experiments described in Section~\ref{sec:cali_setup}, we reveal in Section~\ref{sec:cali_existing} the results that these methods have limitations for prompt selection and even answer selection across numerous datasets. In response, we propose an enhanced calibration method, \underline{C}alibration \underline{B}y \underline{M}arginalization (CBM), in Section~\ref{sec:cali_unified}. Section~\ref{sec:cali_result} shows that CBM notably improves prompt selection for most methods, particularly MI and \mdlm, enabling them to achieve the highest prompt selection performance compared to all other methods. Furthermore, CBM's answer selection enhancement is the most robust across various datasets when compared to existing calibration methods.

\subsection{Experimental Setup for Probability Calibration}\label{sec:cali_setup}

\input{tables/calibration}

We compare the prompt selection performance with four different scenarios of calibration: without applying any calibration; (A) applying calibration only for \underline{A}nswer selection, computing \qq where $\argmax_y$ \qq is selected as the answer; (P) applying calibration only for \underline{P}rompt selection; and (PA) applying calibration for both \underline{P}rompt selection and \underline{A}nswer selection.

Normalization of \qq is not required for answer selection, as it does not affect the $\argmax$ of the scores. However, to obtain PSS, it is essential to normalize \qq so that the sum equals one, thereby preserving the original probabilistic motivation of different methods. Consequently, we apply the softmax function to convert \qq into a proper probability distribution $q(\mathbf{y}|x,t)$.\footnote{To calculate PMI$_\text{DC}$, it is necessary to manually select $x_\text{domain}$ for each prompt in every dataset. Nonetheless, our experiments involve a total of 1,300 unique prompts, making a manual determination of different $x_\text{domain}$ for each prompt a tedious task. Therefore, we use the prompt instantiated with an empty input ($x_\text{domain} = \iota(\text{``''},t)$) for each prompt.}

\subsection{Experimental Results: Underperformance of Existing Calibration Methods}\label{sec:cali_existing}
We check the prompt selection performance of each method across the four calibration scenarios. Surprisingly, for both CC and \pmidc, we find that all three calibration scenarios show degraded performance compared to the scenario of no calibration. Not only does the prompt selection performance degrade, but the best, worst, and average prompt performance also drops in the case of A (only answer selection). This is unexpected, as CC and \pmidc have been reported to improve performance in slightly different setups (our results are in a zero-shot setting, while the main setup of \citet{Zhao2021-lq} is few-shot, and the choice of $x_\text{domain}$ differs for \pmidc).

To further investigate the subpar performance in case A, we analyze the proportion of prompts (out of 100) that exhibit improved performance after applying calibration for answer selection across ten different models and 13 datasets. Figure~\ref{fig:main_answer} displays the average ratio for all models. The figure indicates that the existing calibration methods do not result in better answer selection for the majority of our evaluation datasets. For instance, more than half of the prompts displayed decreased performance after applying CC in 7 out of 13 datasets. A similar pattern holds when applying \pmidc.

\subsection{Enhanced Calibration Method: Calibration By Marginalization (CBM)}\label{sec:cali_unified}
Table~\ref{tab:calibration} shows that the equation for CC can be alternatively expressed as follows:
{\begin{align*}
    \tilde{q}(\mathbf{y}|x,t) = & \text{diag}({\tilde{\mathbf{p}}}_{\text{cf}})^{-1} p(\mathbf{y}|x,t) + \mathbf{0} = \frac{p(\mathbf{y}|x,t)}{{\tilde{\mathbf{p}}}_{\text{cf}}}\\
                       = & \frac{p(\mathbf{y}|x,t)}{\frac{1}{|\mathfrak{C}|} \sum_{\mathfrak{c} \in \mathfrak{C}} \tilde{p}(\mathbf{y}|\mathfrak{c},t)},
\end{align*}
}%
\noindent
which turns CC into a special case of PMI$_\text{DC}$\footnote{We can ignore the lack of $\log$ because it does not change the result of $\argmax$.}, where $\tilde{p}(\mathbf{y}|x_\text{domain},t) = \frac{1}{|\mathfrak{C}|} \sum_{\mathfrak{c} \in \mathfrak{C}} \tilde{p}(\mathbf{y}|\mathfrak{c},t)$. Additionally, upon revisiting the motivation of PMI$_\text{DC}$ and considering the equation of pointwise mutual information $\text{PMI}(x,y) = \log \frac{p(y|x)}{p(y)}$, it becomes evident that $\tilde{p}(\mathbf{y}|x_\text{domain},t)$ approximates ${p}(\mathbf{y}|t)$. Therefore, the distinction between CC and \pmidc lies solely in how they approximate $p(\mathbf{y}|t)$. However, since the approximation for CC relies on three inputs and \pmidc on just one, both methods fall short of providing a stable approximation. This limitation naturally leads to the following question: \textit{Could there be a way to approximate \prob in a more stable manner?}

Encouragingly, the answer to the question is \textit{yes}. A better approximation of \prob can be calculated using the law of marginal probability: $p(\mathbf{y}|t) = \sum_{x \in X} p(\mathbf{y},x|t) = \sum_{x \in X} p(\mathbf{y}|x,t)p(x|t)$. With this more stable approximation of $p(\mathbf{y}|t)$ and the probabilistic assumption of MI that $p(x|t) = \frac{1}{|X|}$, we introduce a new calibration method called \underline{C}alibration \underline{B}y \underline{M}arginalization (CBM) that employs the following equation for answer selection:
{\begin{align}\label{eq:cbm}
\tilde{q}(\mathbf{y}|x,t) = \frac{p(\mathbf{y}|x,t)}{p(\mathbf{y}|t)} = \frac{p(\mathbf{y}|x,t)}{ \frac{1}{|X|} \sum_{x' \in X}p(\mathbf{y}|x',t)}.
\end{align}
}%

Since the calculation of $p(\mathbf{y}|x,t)$ for all $t \in T$ and $x \in X$ is already done to perform prompt selection, CBM does not introduce any additional computational cost for calibration, unlike CC or \pmidc that require inference on additional inputs such as ``N/A'', ``[MASK]'', ``'', and $x_\text{domain}$.

\subsection{Experimental Results: Improvement with CBM Calibration}\label{sec:cali_result}

\begin{figure}[t!]
\centering
\begin{subfigure}[b]{0.48\textwidth}
    \includegraphics[width=\textwidth]{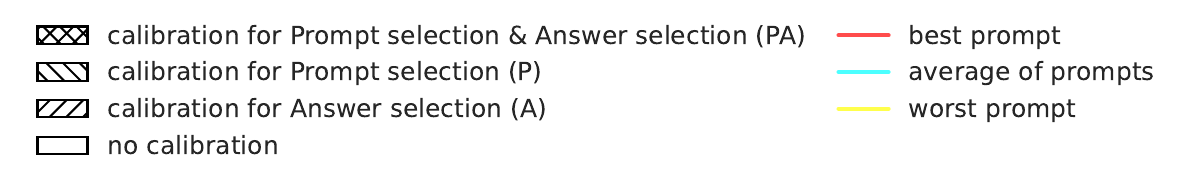}
    \end{subfigure}
    \begin{subfigure}[b]{0.48\textwidth}
    \includegraphics[width=\textwidth]{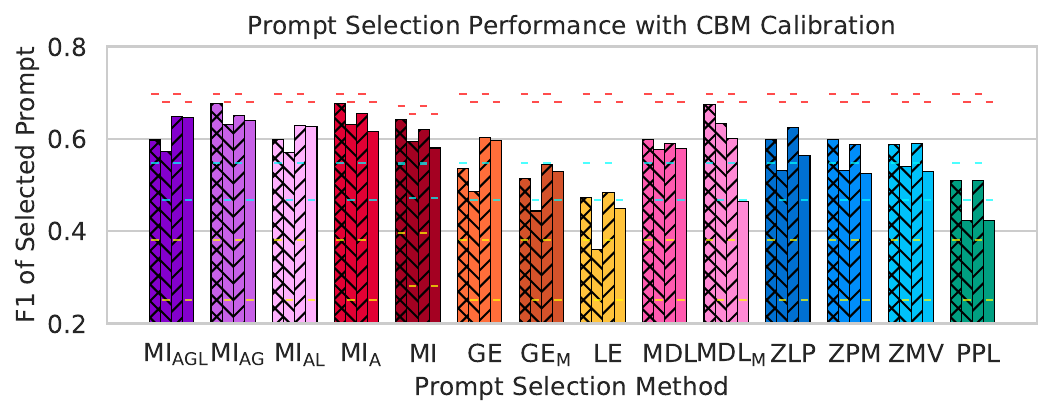}
    \end{subfigure}
\caption{F1 of the prompts selected by different probability-based prompt selection methods, averaged across 13 datasets, for each scenario of CBM calibration.}\label{fig:calibration_ours} 
\end{figure}

\begin{figure*}[t!]
\centering
    \begin{subfigure}[b]{0.534\textwidth}\includegraphics[width=\textwidth]{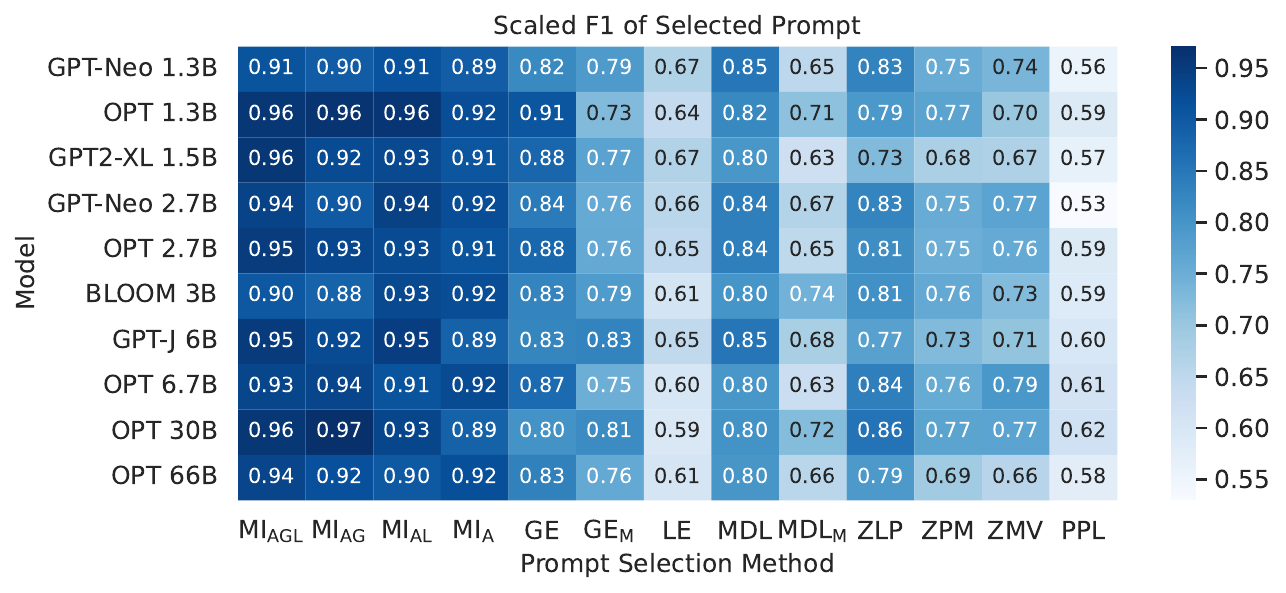}\caption{Scaled F1}\label{fig:model_scaled} \end{subfigure}
    \begin{subfigure}[b]{0.46\textwidth}\includegraphics[width=\textwidth]{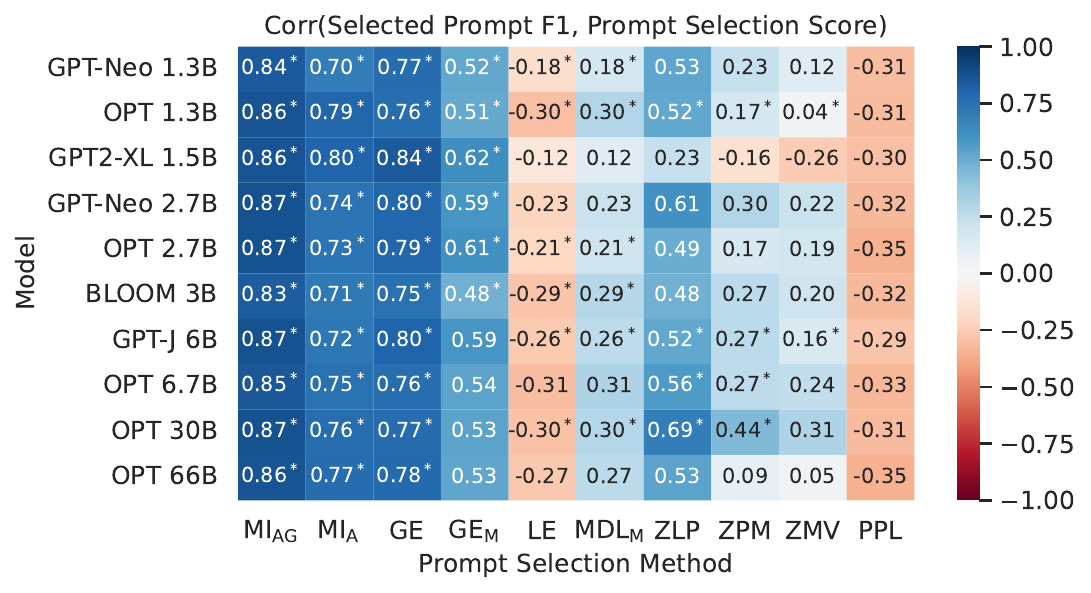}\caption{Correlation of F1 of the selected prompts}\label{fig:model_corr} \end{subfigure}
\caption{Scaled F1 and correlation of F1 of the selected prompts and Prompt Selection Score of different probability-based prompt selection methods for different models, averaged across 13 datasets.}
\end{figure*}
Figure~\ref{fig:calibration_ours} presents the prompt selection performance of each probability-based prompt selection method across the four calibration scenarios of applying CBM. Applying CBM calibration for answer selection (A) enhances prompt selection performance across all methods. Scenarios involving calibration for prompt selection (PA, P) mostly result in unchanged or decreased prompt selection performance compared to the cases without calibration, and applying calibration solely for prompt selection (P) consistently results in diminished performance.

The methods displaying the most significant performance improvements in the PA scenario are \miag, \mia, MI, and \mdlm, particularly with the prompt selection performance of \mic and \lec being the highest among different methods. On average, \mic increases the scaled F1 from 87.79\% (0.5965/0.6795) to 99.44\% (0.6757/0.6795) compared to the best existing method (GE) when the oracle prompt without calibration is used as the target of comparison. The scaled F1 of \mic calculated with respect to the oracle prompt with calibration is 96.85\% (0.6757/0.6977).

Next, we assess the effectiveness of CBM calibration for answer selection by examining the proportion of prompts (out of 100) that show improved performance after applying calibration for answer selection. Figure~\ref{fig:main_answer} indicates that CBM is considerably more effective than CC and \pmidc in enhancing the performance of the prompts. The performance of more than half of the prompts increases after applying CBM in all 13 datasets. Additionally, the performance of nearly 100\% of prompts improves with CBM calibration in 7 datasets. While CC and \pmidc improved almost none of the F1 of the prompts in story and hella, the performance of approximately 70\% of the prompts increased with CBM calibration, possibly due to the more accurate calculation of $p(\mathbf{y}|t)$ as discussed in Section~\ref{sec:cali_unified}.

\section{Discussion}\label{sec:more_analysis}
\input{figtex/discussion}
In this section, we discuss various findings that are relevant to our main experiments.

Figure~\ref{fig:model_scaled} shows that \textit{the effectiveness of a probability-based prompt selection method remains consistent across models of different types and numbers of parameters}, justifying our choice of using a single model (OPT 2.7B) as the representative for all experiments. Figure~\ref{fig:model_corr} shows that the trend of correlation between Prompt Selection Score and performance of the selected prompt is also quite consistent between different models.

Figure~\ref{fig:main_std} shows the mean and standard deviation of the result of prompt selection among five different subsets of 50 prompts randomly sampled from the full set of 100 prompts, using the mainly discussed methods. The result shows that the performance of instance-wise prompt selection methods (\miagl, \mial, MDL) is not stable, likely due to the noisy nature of selecting one prompt for each instance. However, the performance of \mic{} and \lec{} still achieves the highest performance and also shows the lowest standard deviation, proving the effectiveness of CBM.

Through additional analysis, we find that (1) while strong performance in prompt selection does not consistently correlate with Prompt Selection Score, a broadly positive correlation is observed when averaged across most methods; (2) CBM improves the performance of MDL$_\text{M}$ by mitigating overconfidence; (3) MI, GE, and CBM methods face limitations when applied to dynamic datasets with extreme label imbalance; (4) top-performing prompt selection methods from the zero-shot setting, like \mic and \lec, retain their effectiveness in the few-shot setting, further validating their robustness across different conditions.

\section{Related Works}\label{sec:rel_works}

Recent advances in large language models (LLMs) have created the paradigm of prompt-based learning, which gives the benefit that a single pretrained LLM can be used to solve a great number of tasks with task-specific prompts. However, the performance of LLMs can heavily fluctuate according to the choice of prompts~\citep{Zhao2021-lq,Holtzman2021-wj,Lu2022-yp}. To mitigate this issue, prompt engineering attempts to find the prompt that results in the most effective performance on the downstream task~\citep{Liu2021-hv}.

Automatic prompt engineering methods can be largely divided into two groups: the methods that use discrete prompts where the prompts are human-understandable actual text strings, and the methods that optimize continuous prompts where the prompts lie in the embedding space of the model~\citep{li2021prefix,Shin2020-pg}. Probability-based prompt selection methods that we study in this work~\ref{sec:overview} fall into the former group; most of the methods of the latter group require gradient-based training, while probability-based prompt selection does not perform any gradient-based update.

Prompt engineering methods using discrete prompts include prompt paraphrasing, prompt generation, and prompt selection. Among these, prompt paraphrasing or generation approaches can be used together with probability-based selection methods; prompt selection can be performed on the prompts generated through prompt paraphrasing or generation~\citep{Jiang2020-rv,Mishra2022-ca,Gao2021-ej,Wang2022-wk,Prasad2022-qr,Kim2022-ka,Deng2022-bg}.
Among prompt selection methods other than the probability-based approaches, a large portion of the methods are not easily utilizable since they require training an additional model and/or the use of an additional component. \citep{Zhang2022-nk} use reinforcement learning for demonstration selection of in-context learning; \citet{Chang2022-ei} train a scorer and estimator for demonstration selection; \citet{Kumar2021-fn,Xu2022-xi} use a genetic algorithm; \citet{Liu2022-nx,Lyu2022-nn,Rubin2022-jx} use retrieval from a corpus to select the prompts.

On the other hand, probability-based prompt selection offers the advantage of prompt selection requiring \textit{only} the output probabilities of the LLM. While the prerequisite is a set of candidate prompts to select from, this data is relatively small in size and can be easily obtained from the research community~\citep{bach2022promptsource} or via machine generation~\citep{openai2023gpt4}. One limitation of these methods, though, is that one cannot use them for closed-source LLMs that are only available via proprietary LLM APIs that do not provide output probability distributions. Also, when the number of candidate prompts $|T|$ and the size of the dataset used to select the prompt $|X|$ is large, the calculation for prompt selection becomes computationally heavy; using a smaller set $X' \in X$ to choose the prompt for $X$ can be helpful in such a case.\looseness-1

\section{Conclusion}
In this paper, we address the need for a comprehensive evaluation to compare the existing probability-based prompt selection methods, which have been proposed and evaluated under varying conditions and datasets. To achieve this, we introduce a unified evaluation setup to compare these methods, conduct a thorough evaluation, and develop a unified framework of the existing probability-based prompt selection methods. Our analysis within this unified framework has provided insights into the relationship among existing methods, enabling the development of several combinational variants that improve performance. Furthermore, our research on probability calibration has revealed the limitations of existing calibration methods and led to the proposal of an enhanced calibration method, Calibration By Marginalization (CBM). CBM not only significantly improves prompt selection performance but also demonstrates robust answer selection enhancement across multiple datasets. We hope that our unified setup provides a foundation for fair evaluation between various prompt selection methods and that our findings yield deeper insights into probability-based prompt selection.

\section*{Acknowledgements}
The authors would like to extend their sincere gratitude to the anonymous reviewers and Action Editor for their highly detailed and insightful comments and feedback. The authors would also like to thank Sang-Woo Lee for valuable feedback and discussions on the project. This work was partly supported by KT grant (2021, A study on a conversational language model that uses long external text as a prompt, 80\%) and Institute of Information \& communications Technology Planning \& Evaluation (IITP) grant funded by the Korea government (MSIT) (No.2021-0-02068, Artificial Intelligence Innovation Hub, 20\%).

\bibliography{tacl2021}
\bibliographystyle{acl_natbib}

\end{document}

%% file: tables/approaches.tex
\begin{table*}[t!]
\centering
\resizebox{0.9\textwidth}{!}{
\begin{tabular}{@{}lll@{}}
\toprule
\textbf{Existing Method}       & \textbf{Abbr.} & \textbf{Selected Prompt: $\argmax_{t \in T} \cdots$} \\ \midrule
{Mutual Information \citep{Sorensen2022-nm}}  & {MI}       & $ \text{H}\left(\frac{1}{|X|} \sum_x p(\mathbf{y}|x,t)\right) - \frac{1}{|X|} \sum_x \text{H}\left(Y|x,t\right)$ \\ \midrule
Entropy \citep{Lu2022-yp} & & \\
  \quad Global Entropy & GE & $\text{H}\left(\frac{1}{|X|} \sum_x \text{one\ hot}(p(\mathbf{y}|x,t))\right)$                                                \\
\quad Local Entropy   & LE           & $\frac{1}{|X|} \sum_x \text{H}\left(Y|x,t\right)$             \\ \midrule
Minimum Description Length \citep{Wu2022-hr} & MDL & $-\text{H}(Y|x,t)$                      \\ \midrule
Zero-Label Prompt Selection \citep{Liao2022-fe} & & $\sum_x \left[ \mathbbm{1}\left\{ \argmax_y p(\mathbf{y}|x,t) = \argmax_y \mathbf{s}(x,\mathbf{y}) \right\} \right]$ \\
 \quad Log-probability Mean & ZLP & \quad $\mathbf{s}(x,\mathbf{y}) = \frac{1}{|T|} \sum_t \log p(\mathbf{y}|x,t)$          \\
\quad Probability Mean  & ZPM   & \quad $\mathbf{s}(x,\mathbf{y}) = \frac{1}{|T|} \sum_t p(\mathbf{y}|x,t)$               \\
\quad Majority Vote     & ZMV   & \quad $\mathbf{s}(x,\mathbf{y}) = \sum_t \mathbbm{1}\{\argmax_{v \in Y} p(\mathbf{y}|x,t) = v\}$ \\ \midrule
Perplexity \citep{Gonen2022-zf}           & PPL      & $ -\frac{1}{|X|} \sum_x \frac{1}{p(x,t)} $       \\
\bottomrule
\end{tabular}
}
\caption{Summary of the existing probability-based prompt selection methods. Notations used in the equations are explained at the end of Section~\ref{sec:task}.}\label{tab:methods}
\end{table*}

%% file: tables/dataset.tex
\begin{table}[t!]
\setlength{\tabcolsep}{4pt} 
\resizebox{0.48\textwidth}{!}{
\begin{tabular}{@{}lllllccccc@{}}
\toprule
\multirow{2}{*}{\textbf{Dataset}} & \multirow{2}{*}{\textbf{Full Name}} & \multirow{2}{*}{\textbf{Split}} & \multirow{2}{*}[-0.3em]{\parbox{1.7cm}{\textbf{\# Used} \\ \textbf{(\# Orig.)}}} & \multicolumn{1}{c}{\multirow{2}{*}{\textbf{Category}}} & \multicolumn{5}{c}{\textbf{Label Ratio}} \\
\cmidrule{6-10}
 & &  &  & & \textbf{0} & \textbf{1} & \textbf{2} & \textbf{3} & \textbf{4} \\
\midrule
imdb & \href{https://huggingface.co/datasets/imdb/viewer/plain_text/test}{imdb} & test & 1000 (25000) & balanced & 0.51 & 0.49 &  &  &  \\
g-sst2 & \href{https://huggingface.co/datasets/glue/viewer/sst2/validation}{glue-sst2} & valid & 872 & balanced & 0.49 & 0.51 &  &  &  \\
agnews & \href{https://huggingface.co/datasets/ag_news/viewer/default/test}{ag\_news} & test & 1000 (7600) & balanced & 0.27 & 0.25 & 0.25 & 0.24 &  \\
g-rte & \href{https://huggingface.co/datasets/glue/viewer/rte/validation}{glue-rte} & valid & 277 &  balanced & 0.53 & 0.47 &  &  &  \\
newspop & \href{https://huggingface.co/datasets/newspop}{newspop} & train & 1000 (93239) & unbalanced & 0.36 & 0.23 & 0.33 & 0.09 &  \\
t-irony & \href{https://huggingface.co/datasets/tweet_eval/viewer/irony/validation}{tweet\_eval-irony} & valid & 955 & unbalanced & 0.60 & 0.40 &  &  &  \\
t-emo & \href{https://huggingface.co/datasets/tweet_eval/viewer/emotion/validation}{tweet\_eval-emotion} & valid & 374 & unbalanced & 0.39 & 0.25 & 0.09 & 0.27 &  \\
sg-cb & \href{https://huggingface.co/datasets/super_glue/viewer/cb/validation}{super\_glue-cb} & valid & 56 &  unbalanced & 0.41 & 0.50 & 0.09 &  &  \\
sst5 & \href{https://huggingface.co/datasets/SetFit/sst5/viewer/default/test}{SetFit/sst5} & test & 1000 (1101) & unbalanced & 0.13 & 0.29 & 0.18 & 0.23 & 0.18 \\
copa & \href{https://huggingface.co/datasets/super_glue/viewer/copa/validation}{super\_glue-copa} & valid & 100 &  dynamic & 0.55 & 0.45 &  &  &  \\
piqa & \href{https://huggingface.co/datasets/piqa/viewer/plain_text/validation}{piqa} & valid & 1000 (1838) &  dynamic & 0.49 & 0.51 &  &  &  \\
story & \href{https://huggingface.co/datasets/story_cloze/viewer/2016/test}{story\_cloze-2016} & test & 1000 (1871) & dynamic & 0.51 & 0.49 &  &  &  \\
hella & \href{https://huggingface.co/datasets/Rowan/hellaswag/viewer/default/validation}{Rowan/hellaswag} & valid & 1000 (10003)  & dynamic & 0.22 & 0.25 & 0.26 & 0.26 &  \\
\bottomrule
\end{tabular}
}
\caption{Datasets chosen to evaluate various probability-based prompt selection methods.}\label{tab:dataset}
\end{table}

%% file: tables/calibration.tex
\begin{table}[t!]
\centering \setlength\extrarowheight{2pt}
\resizebox{0.47\textwidth}{!}{
\begin{tabular}{ll}
\toprule
\textbf{Existing Method} & \textbf{Equation for Answer Selection} \\
\midrule
\multirow{4}{*}{\parbox{3cm}{Contextual \\Calibration (CC)~\citep{Zhao2021-lq}}} & $\mathfrak{C} = \{ \text{"N/A"}, \text{"[MASK]"}, \text{""} \}$ \\
 & ${\tilde{\mathbf{p}}}_{\text{cf}} = \frac{1}{|\mathfrak{C}|} \sum_{\mathfrak{c} \in \mathfrak{C}} \tilde{p}(\mathbf{y}|\mathfrak{c},t)$ \\
 & $\mathbf{W} = \text{diag}(\tilde{\mathbf{p}}_{\text{cf}})^{-1}, \mathbf{b} = \mathbf{0}$ \\
 & $\tilde{q}(\mathbf{y}|x,t) = \mathbf{W} p(\mathbf{y}|x,t) + \mathbf{b} $ \\
 \gcmidrule(lr){1-2}
\multirow{4}{*}{\parbox{3cm}{Domain \\Conditional PMI \\(PMI$_\text{DC}$)~\citep{Holtzman2021-wj}}} & \multirow{4}{*}{$\tilde{q}(\mathbf{y}|x,t) = \log\dfrac{\tilde{p}(\mathbf{y}|x,t)}{\tilde{p}(\mathbf{y}|x_\text{domain},t)}$} \\
 & \\
 & \\
 & \\
\bottomrule
\end{tabular}
}
\caption{Existing calibration methods proposed for answer selection. $\argmax_y \tilde{q}(\mathbf{y}|x,t)$ is selected as the answer for the prompt $t$ instantiated by input instance $x$. Note that the actual calculation of CC in the official code uses $\mathbf{p}_\text{cf}$, mean-normalized $\tilde{\mathbf{p}}_\text{cf}$; thus, we also use it in our experiments.}\label{tab:calibration}
\end{table}

%% file: figtex/discussion.tex
\begin{figure*}[ht!]
\centering
    \begin{minipage}[b]{0.19\textwidth}
    \includegraphics[width=\textwidth]{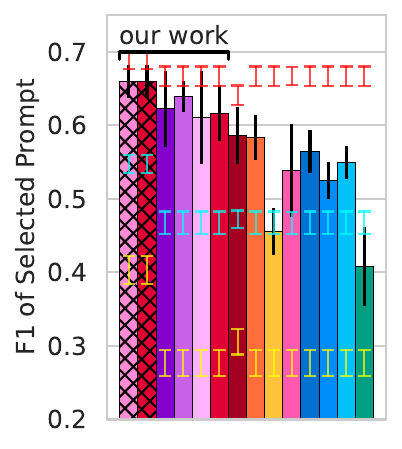}
    \caption{Mean and standard deviation of prompt selection among five sets of 50 prompts, sampled from the full set of 100 prompts.}
    \label{fig:main_std}
    \end{minipage}\hfill
    \begin{minipage}[b]{0.78\textwidth}
    \begin{subfigure}[t]{0.99\textwidth}
    \includegraphics[width=\textwidth]{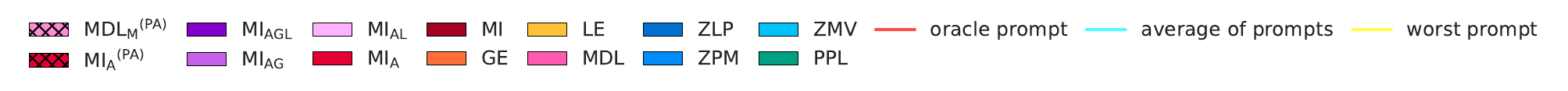}
    \end{subfigure}
        \begin{subfigure}[t]{0.99\textwidth}
        \includegraphics[width=\textwidth]{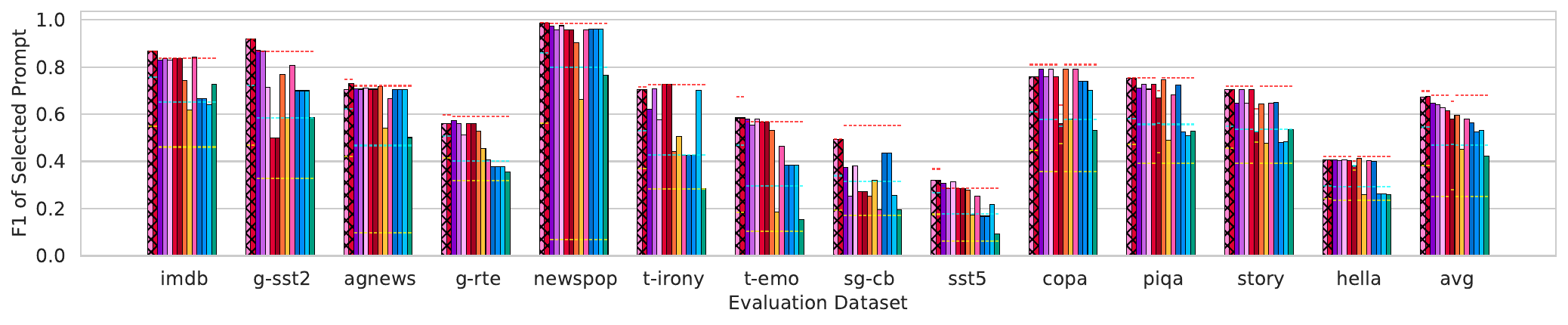}
        \end{subfigure}
        \begin{subfigure}[t]{0.99\textwidth}
        \includegraphics[width=\textwidth]{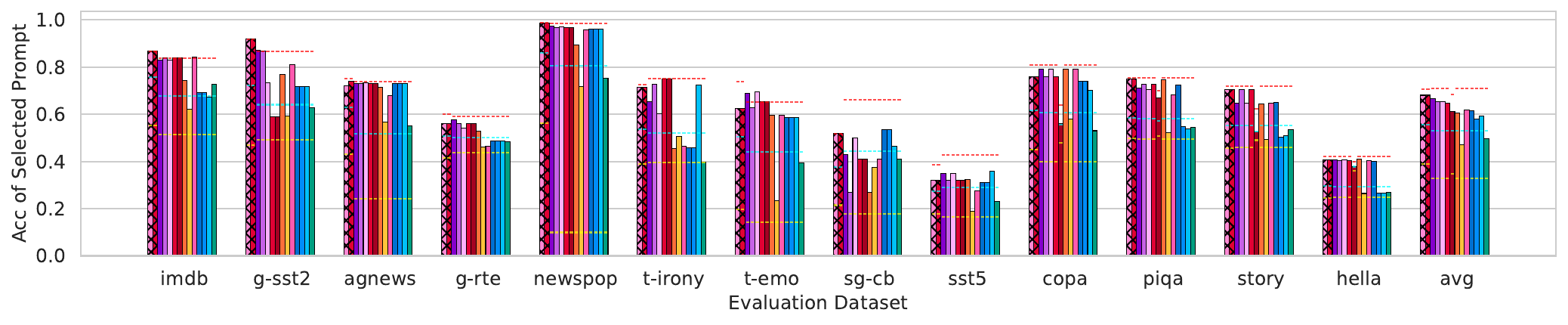}
        \end{subfigure}
    \caption{F1 (top) and accuracy (bottom) of the prompts selected by the different probability-based prompt selection methods, shown for each dataset.}
    \label{fig:granular}
    \end{minipage}
\label{fig:discussion}
\end{figure*}